\documentclass[letterpaper]{article} 
\usepackage{aaai25}  
\usepackage{times}  
\usepackage{helvet}  
\usepackage{courier}  
\usepackage[hyphens]{url}  
\usepackage{graphicx} 
\usepackage{natbib}  
\usepackage{caption} 
\usepackage{algorithm}
\usepackage{algorithmic}

\usepackage{amsmath}
\usepackage{booktabs} 
\usepackage{multirow}
\usepackage{amssymb}
\newtheorem{definition}{Definition}
\usepackage{amsfonts}
\usepackage{nicefrac} 
\usepackage[table,xcdraw]{xcolor}
\usepackage{bm}

\usepackage{newfloat}
\usepackage{listings}
\DeclareCaptionStyle{ruled}{labelfont=normalfont,labelsep=colon,strut=off} 
\floatstyle{ruled}
\newfloat{listing}{tb}{lst}{}
\floatname{listing}{Listing}
\title{Pareto Continual Learning: Preference-Conditioned Learning and Adaption for Dynamic Stability-Plasticity Trade-off}
\author{
    Song Lai\textsuperscript{\rm 1,2,5},
    Zhe Zhao\textsuperscript{\rm 1,6},
    Fei Zhu\textsuperscript{\rm 2},
    Xi Lin\textsuperscript{\rm 1},
    Qingfu Zhang\textsuperscript{\rm 1,5}\thanks{Corresponding authors.},
    Gaofeng Meng\textsuperscript{\rm 2,3,4}\footnotemark[1]
}
\affiliations{
    \textsuperscript{\rm 1} Department of Computer Science, City University of Hong Kong\\
    \textsuperscript{\rm 2} Centre for Artificial Intelligence and Robotics, HK Institute of Science \& Innovation, Chinese Academy of Sciences\\
    \textsuperscript{\rm 3} State Key Laboratory of Multimodal Artificial Intelligence Systems, Institute of
Automation, Chinese Academy of Sciences\\
    \textsuperscript{\rm 4} School of Artificial Intelligence, University of Chinese Academy of Sciences \\
    \textsuperscript{\rm 5} City University of Hong Kong Shenzhen Research Institute \\
    \textsuperscript{\rm 6} University of Science and Technology of China \\
    \{songlai2-c, xi.lin\}@my.cityu.edu.hk, zz4543@mail.ustc.edu.cn\\
    qingfu.zhang@cityu.edu.hk, \{fei.zhu, gaofeng.meng\}@cair-cas.org.hk 
}

\usepackage{bibentry}

\begin{document}

\maketitle

\begin{abstract}
Continual learning aims to learn multiple tasks sequentially. A key challenge in continual learning is balancing between two objectives: retaining knowledge from old tasks (stability) and adapting to new tasks (plasticity). Experience replay methods, which store and replay past data alongside new data, have become a widely adopted approach to mitigate catastrophic forgetting. However, these methods neglect the dynamic nature of the stability-plasticity trade-off and aim to find a fixed and unchanging balance, resulting in suboptimal adaptation during training and inference. In this paper, we propose Pareto Continual Learning (ParetoCL), a novel framework that reformulates the stability-plasticity trade-off in continual learning as a multi-objective optimization (MOO) problem. ParetoCL introduces a preference-conditioned model to efficiently learn a set of Pareto optimal solutions representing different trade-offs and enables dynamic adaptation during inference. From a generalization perspective, ParetoCL can be seen as an objective augmentation approach that learns from different objective combinations of stability and plasticity. Extensive experiments across multiple datasets and settings demonstrate that ParetoCL outperforms state-of-the-art methods and adapts to diverse continual learning scenarios.  
\end{abstract}

%
\begin{links}
    \link{Code}{https://github.com/laisong-22004009/ParetoCL}
\end{links}

\section{Introduction}

Continual learning (CL) has emerged as a crucial paradigm, enabling artificial intelligence systems to learn and adapt in dynamic environments. Unlike traditional machine learning~\cite{lecun2015deep,he2016deep}, where models are trained on a fixed dataset, continual learning enables models to learn from a stream of tasks or data distributions sequentially. This ability is crucial for real-world applications with incremental data and evolving environments. However, a fundamental challenge in continual learning is the stability-plasticity dilemma~\cite{wang2024comprehensive}, which refers to the trade-off between a model's ability to retain learned knowledge (stability) and its ability to acquire new knowledge (plasticity).

Numerous CL methods have been proposed to tackle the stability-plasticity dilemma. Among them, experience replay (ER) methods~\cite{Rolnick2018ExperienceRF,buzzega2020dark,arani2021learning} stand as a straightforward yet effective research line. These methods store a subset of past data in a memory buffer and replay it alongside new data during training. However, existing ER methods aim to find a fixed trade-off between stability and plasticity, ignoring the dynamic nature of the stability-plasticity trade-off. This results in a single model that cannot dynamically adapt to evolving environments and changing task requirements during training. Second, the optimal stability-plasticity trade-off may vary not only during training but also during inference. For example, in the study by~\cite{cai2021online}, it was observed that the stability-plasticity trade-off evolves over time as the data distribution shifts. This highlights the need for dynamic adaptation of the trade-off based on the current learning context. 

\textit{Can we make a proper trade-off between stability and plasticity and integrate these two objectives to enhance the model's generalization ability?} To address this question, we draw inspiration from multi-objective optimization (MOO)~\cite{miettinen1999nonlinear}. MOO aims to find a set of Pareto optimal solutions that represent different optimal trade-offs among conflicting objectives. Building upon this concept, we envision finding a set of Pareto optimal models that embody different stability-plasticity trade-offs rather than a single model with a fixed trade-off. However, training separate models for each trade-off is computationally expensive, and a finite number of Pareto solutions cannot fully capture the relationship between stability and plasticity. To overcome these challenges, we propose Pareto Continual Learning (ParetoCL), a novel framework that introduces an efficient preference-conditioned model to \textit{learn} the trade-off between the two objectives and perform dynamic adaptation. Specifically, ParetoCL trains the preference-conditioned model by sampling preference vectors from a prior distribution and minimizes a preference-based aggregation of the losses on the replay buffer (for stability) and the new data (for plasticity). During inference, the model dynamically selects the optimal trade-off for each sample based on the preference vector that yields the most confident prediction. Experiments on multiple benchmarks demonstrate that ParetoCL significantly outperforms state-of-the-art continual learning methods. By inputting different preferences, ParetoCL can obtain a set of well-distributed Pareto optimal solutions with varying trade-offs between stability and plasticity.

From a generalization perspective, our work can be seen as an extension of data augmentation~\cite{buzzega2020dark} and class augmentation~\cite{zhu2021prototype} in the continual learning domain—\textit{objective augmentation}. The model is trained using different combinations of preferences between  stability and plasticity, equivalent to learning from different combinations of old and new task distribution. Since we cannot know which task distribution combination leads to better generalization in advance, and training models for each possible distribution is prohibitively expensive, we transform the problem of combining task distribution into that of combining preference-conditioned model hypotheses. In this way, we bypass suboptimal task combinations and promote positive transfer between past and current knowledge.

The main contributions of this work are threefold:
\begin{itemize}
\item \textit{New perspective}: To the best of our knowledge, this work is the first attempt to explore the dynamic nature of the stability-plasticity trade-off in continual learning and reformulate it as a MOO problem.
\item \textit{Effective algorithm}: We propose an efficient preference-conditioned CL model that learns a set of Pareto solutions of the stability-plasticity trade-off during training and enables dynamic adaptation during inference.
\item \textit{Compelling empirical results}: We conduct extensive experiments on multiple benchmarks and settings, demonstrating that ParetoCL significantly outperforms state-of-the-art continual learning methods.
\end{itemize}

\section{Related Work}
\subsection{Continual Learning}
Continual learning methods can be broadly categorized into regularization-based, expansion-based, and replay-based approaches. Regularization-based methods~\cite{Huszr2017NoteOT,Zenke2017ContinualLT,yang2019adaptive,yang2021cost} introduce additional regularization terms to the loss function to prevent catastrophic forgetting. These regularizers penalize changes to parameters that are important for previous tasks. Expansion-based methods~\cite{li2019learn,serra2018overcoming,yoon2017lifelong,mallya2018piggyback,hung2019compacting} allocate new resources for new tasks while preserving the learned knowledge in the existing network. Replay-based methods~\cite{Rolnick2018ExperienceRF, buzzega2020dark,aljundi2019gradient} store and replay previous samples to maintain learned information, which has been widely adopted due to its effectiveness and simplicity. Methods like iCaRL~\cite{rebuffi2017icarl} combine replay with knowledge distillation to retain previous knowledge. MER~\cite{riemer2018learning} and CLSER~\cite{arani2021learning} introduce additional model parameters to consolidate task-specific knowledge. ER-ACE~\cite{Caccia2021NewIO} modifies the loss function to alleviate representation drift. However, these approaches seek a fixed trade-off between stability and plasticity, limiting their ability to adapt to dynamic learning scenarios.

\subsection{Multi-Objective Optimization}
Multi-objective optimization (MOO) aims to find a set of Pareto optimal solutions representing different trade-offs among multiple conflicting objectives~\cite{ehrgott2005multicriteria}. MOO has found numerous applications in machine learning, including reinforcement learning~\cite{liu2014multiobjective}, neural architecture search~\cite{lu2019nsga}, and fairness~\cite{wu2022multi}. Gradient-based MOO algorithms, such as MDGA~\cite{sener2018multi} and ParetoMTL~\cite{lin2019pareto}, have been proposed to train large-scale neural networks for multi-task learning. These methods scalarize the MOO problem into a single-objective problem using a weighted sum of the objectives. However, they require a priori preference specification and can only find a single Pareto optimal solution. Recent works~\cite{lin2022pareto,navon2020learning} propose to approximate the entire Pareto front by training a hypernetwork conditioned on the preference vector. Our work shares the goal of approximating the Pareto front of multiple objectives with these MOO methods. However, we focus on the specific problem of continual learning and propose a novel preference-conditioned model to adapt the stability-plasticity trade-off dynamically during both training and inference.

\section{Preliminary}
\subsection{Experience Replay}
Continual learning requires the model to learn from a non-stationary data stream. At each time step $t$, the model receives a batch of labeled data $\mathcal{D}_t = \{(x_i^t, y_i^t)\}_{i=1}^{B}$ drawn from an underlying distribution $\mathcal{P}_t$ that gradually shifts over time. The goal is to learn a model $f_{\theta}$ parameterized by $\theta$ that performs well on all seen distributions $\{\mathcal{P}_1, ..., \mathcal{P}_t\}$ up to the current time step.

Let $\mathcal{L}(f_\theta; \mathcal{D})$ denote the loss of the model $f_\theta$ on a dataset $\mathcal{D}$. To tackle catastrophic forgetting, experience replay methods maintain a memory buffer $\mathcal{M}$ that stores a subset of past data. At each time step $t$, the model is updated using the following loss:
\begin{equation}
\mathcal{L}_t = \lambda \mathcal{L}_{new}(f_\theta; \mathcal{D}_t) + (1-\lambda) \mathcal{L}_{replay}(f_\theta; \mathcal{M}_t),
\end{equation}
where $\lambda \in [0, 1]$ is a hyperparameter that controls the trade-off between the losses on the new data $\mathcal{D}_t$ and the memory buffer $\mathcal{M}_t$. The first term promotes plasticity to adapt to the new data, while the second term promotes stability from past data. The memory buffer $\mathcal{M}_t$ is updated by randomly selecting samples to replace with the new data.

\begin{figure*}[t]
    \centering
    \includegraphics[width=1\textwidth]{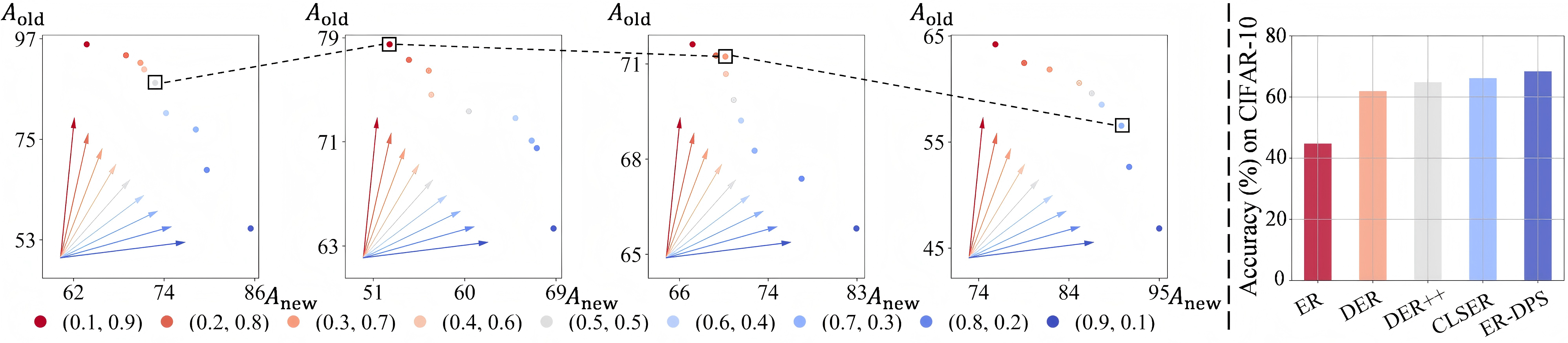}
    \caption{ \textbf{Motivation.} (left) illustrates the multi-stage training process of a ResNet-18 model on 5 tasks of Seq-CIFAR10 while (right) compares ER-DPS with several SOTA replay-based methods. In (left), each stage builds upon the model trained on the previous task, incorporating different stability-plasticity trade-offs. Each point in (left) corresponds to the accuracy obtained when training the model under a specific preference vector, demonstrating how different trade-offs influence stability and plasticity. The metrics in the graph, \(A_{\text{old}}\) and \(A_{\text{new}}\), represent the average accuracy on previous tasks and the accuracy on the current task, respectively, corresponding to stability and plasticity. The dashed line indicates the model selected at the current stage corresponding to the chosen preference. Detailed data is provided in supplementary material.} \label{fig:multistage}
\end{figure*}

\subsection{Multi-Objective Optimization}
Multi-objective optimization (MOO) problem is defined as:
\begin{equation}
\min_{\theta} \mathcal{F}(\theta) = (f_1(\theta), ..., f_m(\theta))^\top \quad \text{s.t.} \quad \theta \in \Theta,
\end{equation}
where $\theta$ is the decision variable, $\Theta$ is the feasible set, and $f_1(\theta), ..., f_m(\theta)$ are $m$ objective functions. Due to the conflicting nature of objectives, no single solution $\theta^*$ can optimize all objectives simultaneously. This leads to the definitions of Pareto dominance, Pareto optimality, and Pareto Set/Front for MOO problems:
\begin{definition}[Pareto Dominance]
A solution $\theta_1$ dominates another solution $\theta_2$, denoted $\theta_1 \prec \theta_2$, if $f_i(\theta_1) \leq f_i(\theta_2)$ for all $i \in \{1, \ldots, m\}$ and there exists $j \in \{1, \ldots, m\}$ such that $f_j(\theta_1) < f_j(\theta_2)$.
\end{definition}

\begin{definition}[Pareto Optimality]
A solution $\theta$ is Pareto optimal if there does not exist another solution $\theta' \in \Theta$ that dominates $\theta$, i.e., $\nexists \theta' \in \Theta$ such that $\theta' \prec \theta$.
\end{definition}

\begin{definition}[Pareto Set and Pareto Front]
The Pareto Set, $\Theta^*$, is the set of all Pareto optimal solutions, defined as $\Theta^* = \{\theta \in \Theta \mid \nexists \theta' \in \Theta \text{ such that } \theta' \prec \theta\}$. The Pareto Front, $\mathcal{F}(\Theta^*)$, is the image of the Pareto Set in the objective space, defined as $\mathcal{F}(\Theta^*) = \{\bm{F}(\theta) \mid \theta \in \Theta^*\}$.
\end{definition}

\section{Pareto Continual Learning}
\subsection{Motivation}
To illustrate the motivation behind our algorithm, we conduct an experiment on class incremental learning with Seq-CIFAR10~\cite{buzzega2020dark}. The goal is to demonstrate the effectiveness of training a diverse set of solutions with different stability-plasticity trade-offs and dynamically selecting the appropriate solution during inference. Figure~\ref{fig:multistage}(left) illustrates the multi-stage training process of a ResNet-18 model on 5 tasks of Seq-CIFAR10. Each stage builds upon the model trained on the previous task and incorporates different stability-plasticity trade-offs by varying the preference vectors in the weighted aggregation of the replay loss and the cross-entropy loss on the current task. The models are evaluated on the test sets of both the current and previous tasks, and the model with the highest average accuracy is selected for the next stage. We call this method ER-DPS (Experience Replay with Dynamic Preference Selection), which can be considered as an \textbf{oracle} ER model with optimal stability-plasticity trade-off in task level. Figure~\ref{fig:multistage}(right) shows the final mean accuracy of the models across all tasks, demonstrating the superiority of dynamic trade-off adaption over fixed weighted aggregation. The experimental results lead to two key observations: 
\begin{itemize}
     \item The optimal stability-plasticity trade-off varies and the model needs to prioritize plasticity in certain stages while emphasizing stability in others;
    \item Selecting the appropriate trade-off carefully improves the performance compared to using a fixed trade-off.
\end{itemize}
However, directly applying the above approach to continual learning faces two challenges:
\begin{itemize}
    \item The optimal trade-off is an unknown priori.
    \item Training separate models for each trade-off is computationally expensive and memory-inefficient.
\end{itemize}
These challenges motivate the design of our Pareto Continual Learning framework, which learns a single model conditioned on preference vectors to approximate the Pareto front of stability-plasticity trade-offs.

\subsection{ER as Multi-Objective Optimization}
Drawing insights from the above observations, we propose to formulate experience replay as a MOO problem. Specifically, let $\mathcal{L}_{new}$ and $\mathcal{L}_{replay}$ denote the losses on the new data and the replay buffer, respectively. We define two objectives corresponding to plasticity and stability:
\begin{equation}
\min_{\theta} F(\theta) = (\mathcal{L}_{replay}(f_{\theta}), \mathcal{L}_{new}(f_{\theta}))^\top.
\end{equation}
Our goal is to find a Pareto set that represent different trade-offs between these two objectives, rather than a single solution with fixed trade-off. This approach enables the model to adapt dynamically to varying degrees of knowledge retention and acquisition across different tasks and time periods.

\begin{figure*}[ht]
\centering
\includegraphics[width=0.95\textwidth]{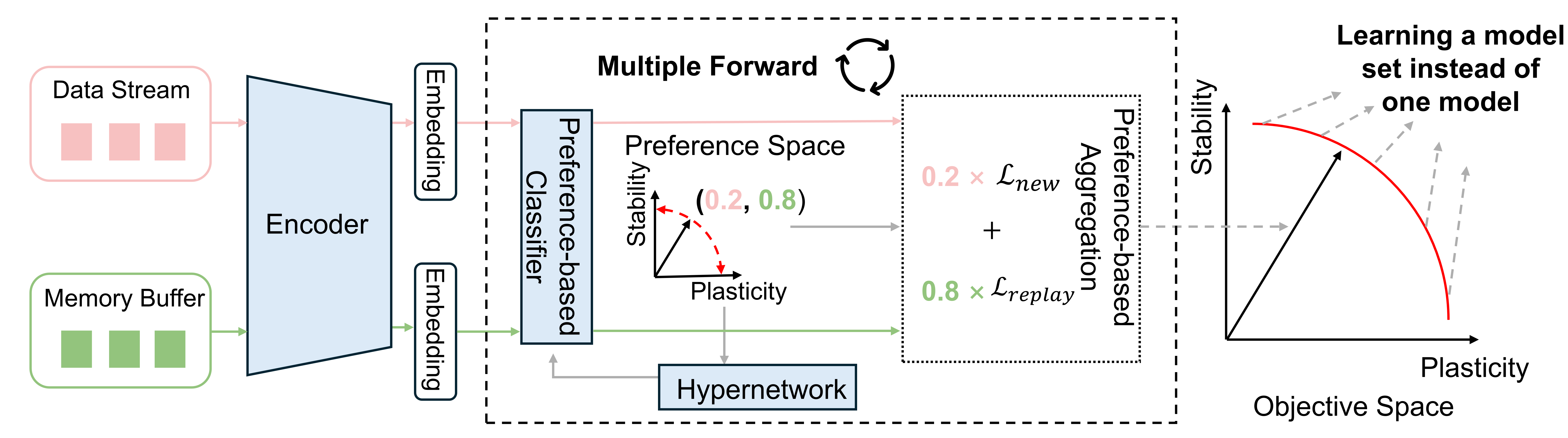}
\caption{\textbf{An overview of our proposed ParetoCL framework}. Our model consists of two parts: a shared encoder and a preference-based classifier (with parameters generated by a hypernetwork). The model takes inputs from the data stream and memory buffer, which are then transformed into embeddings by the encoder. These embeddings are used by the preference-based classifier multiple times with different preferences, each of which is sampled from a prior distribution. The sampled preference is then fed into the hypernetwork to obtain the parameters of the preference-based classifier. For the embeddings from both the data stream and memory buffer, the classifier computes two losses: $\mathcal{L}_{new}$ and $\mathcal{L}_{replay}$. The final optimization objective of the model is the expectation of the preference-based aggregation of these two losses. Through this approach, ParetoCL learns a mapping from the preference space to the objective space of different plasticity-stability trade-offs.} \label{fig:overview}
\end{figure*}

\begin{algorithm}[!ht]
\caption{ParetoCL Training}
\label{alg:paretocl_training}
\begin{algorithmic}[1]
\REQUIRE Prior distribution of preferences $p(\alpha)$, model parameter $\theta$
\STATE Initialize $\theta$
\FOR{each iteration}
\STATE Sample data $\mathcal{D}_t$ from the data stream
\STATE Sample data $\mathcal{M}_t$ from the memory buffer
\STATE Feature embedding: $h_{new} = f(\theta, \mathcal{D}_t)$ 
\STATE Feature embedding: $h_{replay} = f(\theta, \mathcal{M}_t)$ 
\FOR{$k = 1$ to $K$}
\STATE Sample preference vector: $\alpha_k \sim p(\alpha)$
\STATE Compute output based on Eq.(\ref{eq:clf})
\ENDFOR
\STATE Compute loss based on Eq.(\ref{eq:loss})
\STATE Update $\theta$
\ENDFOR
\end{algorithmic}
\end{algorithm}

\begin{algorithm}[!ht]
\caption{ParetoCL Inference}
\label{alg:paretocl_inference}
\begin{algorithmic}[1]
\REQUIRE Model parameter $\theta$
\STATE Input sample: $x$
\STATE Feature embedding: $h = f(\theta, x)$
\FOR{$k = 1$ to $K$}
\STATE Sample preference vector $\alpha^k \sim p(\alpha)$
\STATE Compute each output based on Eq.(\ref{eq:clf})
\ENDFOR
\STATE Compute final output based on Eq.(\ref{eq:final})
\end{algorithmic}
\end{algorithm}

\subsection{Preference-Conditioned Learning}

To learn a set of models with different trade-offs, we introduce a preference-conditioned model $f_\theta(x; \alpha)$ parameterized by $\theta$ that takes as input a sample $x$ and a preference vector $\alpha \in \mathbb{R}^2_+$ and outputs a prediction, as illustrated in Figure~\ref{fig:overview}. The preference vector $\alpha = (\alpha_1, \alpha_2)$ determines the trade-off between stability and plasticity, where a higher value of $\alpha_1$ prioritizes stability and a higher value of $\alpha_2$ prioritizes plasticity. The model is trained to minimize the following loss:
\begin{equation}
\mathcal{L}(\theta) = \mathbb{E}_{\alpha \sim p(\alpha)} [\alpha_1 \mathcal{L}_{replay}(\theta, \alpha) + \alpha_2 \mathcal{L}_{new}(\theta, \alpha)],
\label{eq:loss}
\end{equation}
where $p(\alpha)$ is a prior distribution over the preferences (e.g., a Dirichlet distribution).

To implement preference conditioning, we employ a hypernetwork $\Psi$ that generates the parameters of the model's final linear layer based on the preference vector. Specifically, let $h_\theta(x)$ denote the penultimate features of the model. The final output is computed as:
\begin{equation}
f_\theta(x; \alpha) = W(\alpha) h_\theta(x) + b(\alpha),
\label{eq:clf}
\end{equation}
where $W(\alpha) = \Psi_W(\alpha)$ and $b(\alpha) = \Psi_b(\alpha)$ are the weight matrix and bias vector generated by the hypernetwork $\Psi$ conditioned on the preference vector $\alpha$.

During training, we sample a new preference vector $\alpha \sim p(\alpha)$ for each mini-batch and compute the losses $\mathcal{L}_{replay}(\theta, \alpha)$ and $\mathcal{L}_{new}(\theta, \alpha)$ using the preference-conditioned model $f_\theta(x; \alpha)$, which is outlined in Algorithm \ref{alg:paretocl_training}. The model parameters $\theta$ and the hypernetwork parameters are jointly optimized to minimize the expected loss over the preference distribution, as shown in Eq.(4). By varying the preference vectors, the model learns to map different preferences to different points on the Pareto front of stability-plasticity trade-offs.

\begin{table*}[]
\centering
\resizebox{\textwidth}{!}
{
\begin{tabular}{@{}l|cccc|cccc|cccc@{}}
\toprule
                         & \multicolumn{4}{c|}{Seq-CIFAR10}                       & \multicolumn{4}{c|}{Seq-CIFAR100}                      & \multicolumn{4}{c}{Seq-TinyImageNet}                  \\ \cline{2-13}
                         & \multicolumn{2}{c}{Online}      & \multicolumn{2}{c|}{Offline}    & \multicolumn{2}{c}{Online}      & \multicolumn{2}{c|}{Offline}    & \multicolumn{2}{c}{Online}      & \multicolumn{2}{c}{Offline}     \\ \cline{2-13}
\multirow{-3}{*}{\textbf{Method}} & AAA            & Acc            & AAA            & Acc            & AAA            & Acc            & AAA            & Acc            & AAA            & Acc            & AAA            & Acc            \\ \midrule
SGD                      & 38.12          & 18.33          & 43.11          & 19.27          & 11.63          & 5.69           & 19.22          & 8.46           & 11.35          & 4.02           & 17.60          & 6.91           \\ \midrule
OnEWC~\cite{Huszr2017NoteOT}                    & 38.75          & 17.89          & 43.25          & 21.33          & 12.14          & 6.23           & 17.84          & 6.64           & 11.92          & 4.73           & 14.64          & 4.92           \\
LwF~\cite{Li2016LearningWF}                      & 35.61          & 18.91          & -           & -           & 12.16          & 5.99           & -         & -          & 11.42          & 4.78           & -         & -         \\
IS~\cite{Zenke2017ContinualLT}                       & 38.89          & 20.44          & 40.52          & 18.78          & 13.21          & 7.33           & 16.56          & 5.25           & 10.82          & 4.23           & 11.42          & 2.83           \\ \midrule
La-MAML~\cite{Gupta2020LaMAMLLM}                  & 36.76          & 35.44          & 61.56          & 44.38          & 13.45          & 12.21          & 32.35          & 20.34          & 16.32          & 13.83          & 37.11          & 24.91          \\
VR-MCL~\cite{wu2024meta}                  & 69.57          & 58.43          & 77.95          & 68.14          & 30.46          & 22.31          & 42.36          & 30.09          & 24.41          & 20.55          & 41.13          & 28.02          \\ \midrule
ER~\cite{Rolnick2018ExperienceRF}                      & 52.74          & 33.14          & 75.24          & 64.13          & 18.70          & 15.12          & 36.73          & 22.19          & 19.14          & 14.21          & 37.87          & 25.95          \\
A-GEM~\cite{Chaudhry2018EfficientLL}                   & 39.25          & 18.54          & 43.63          & 20.62          & 14.66          & 6.24           & 19.89          & 7.94           & 11.66          & 5.26           & 19.23          & 8.40           \\
GEM~\cite{LopezPaz2017GradientEM}                     & 36.83          & 19.75          & 49.68          & 24.45          & 16.23          & 9.91           & 22.35          & 11.45          & 13.57          & 6.89           & 20.32          & 9.28           \\
DER~\cite{buzzega2020dark}                      & 38.59          & 18.10          & 66.48          & 59.55          & 14.65          & 6.27           & 19.22          & 7.85           & 11.58          & 5.31           & 17.71          & 7.18           \\
DER++~\cite{buzzega2020dark}                    & 60.63          & 50.33          & 75.20          & 67.36          & 24.47          & 16.32          & 40.01          & 23.86          & 19.21          & 13.65          & 38.94          & 27.07          \\
CLSER~\cite{arani2021learning}                  & 61.88          & 49.03          & 76.25          & 67.86          & 27.83          & 19.45          & 41.28          & 27.06          & 23.47          & 19.72          & 38.64          & 26.48          \\
ER-OBC~\cite{Chrysakis2023OnlineBC}                  & 65.77          & 54.64          & 76.42          & 65.93          & 26.11          & 17.31          & 42.26          & \textbf{30.22} & 24.11          & 19.86          & 39.41          & 27.67          \\
OCM~\cite{Guo2022OnlineCL}                     & 66.22          & 53.89          & -           & -          & 26.03          & 15.88          & -           & -           & 17.55          & 8.33           & -           & -           \\
\rowcolor[HTML]{EFEFEF} 
\textbf{ParetoCL (Ours)}          & \textbf{70.89} & \textbf{59.95} & \textbf{78.98} & \textbf{69.55} & \textbf{33.04} & \textbf{24.45} & \textbf{44.32} & 28.79          & \textbf{31.72} & \textbf{23.09} & \textbf{43.02} & \textbf{28.28} \\ \bottomrule
\end{tabular}
}
\caption{Performance of continual learning in Seq-CIFAR10 and longer task sequences SeqCIFAR100, Seq-TinyImageNet with ResNet-18 under online and offline settings. ‘-’ indicates the implementation is unstable. All reported results are the average of 5 runs. Full table can be found in supplementary material.}\label{tab:baseline_comp}
\end{table*}

\subsection{Dynamic Preference Adaptation}
With the preference-conditioned model, we can adapt the stability-plasticity trade-off on the fly during inference. As shown in Algorithm \ref{alg:paretocl_inference}, for each input sample $x$, we sample multiple preference vectors $\{\alpha^1, ..., \alpha^K\}$ and compute the model outputs $\{f_\theta(x; \alpha^1), ..., f_\theta(x; \alpha^K)\}$. We then select the output with the highest confidence (i.e., lowest entropy) as the final prediction:

\begin{equation}
    y^* = \arg\min_{k} H(f_\theta(x; \alpha^k)),
\label{eq:final}
\end{equation}
where $H(\cdot)$ denotes the entropy of the model's output distribution. Intuitively, the output with the lowest entropy corresponds to the preference vector that yields the most confident prediction for the given sample. This dynamic preference adaptation allows the model to select the most suitable trade-off for each sample, leading to better performance than using a fixed trade-off.

\subsection{Comparison with Existing Work}
While POCL~\cite{Wu2024MitigatingCF} also incorporates Pareto optimization into CL, ParetoCL differs in several aspects: First, ParetoCL explicitly models the stability-plasticity trade-off as a MOO problem, while POCL focuses on modeling relationships among past tasks to enhance stability; Second, instead of using gradient alignment like POCL, ParetoCL learns a set of Pareto solutions, which allows ParetoCL to handle the dynamic nature of the trade-off.

\section{Experiments} 
\subsection{Experimental Setup}
\noindent\textbf{Datasets and Settings}
To verify the effectiveness of the proposed ParetoCL, we conduct comprehensive experiments on commonly used datasets Seq-CIFAR10, as well as the longer task sequences Seq-CIFAR100 and Seq-TinyImageNet~\cite{buzzega2020dark}. Specifically, the Seq-CIFAR10 dataset comprises 5 tasks, with each task containing 2 classes. Seq-CIFAR100 consists of 10 tasks, each with 10 classes, while Seq-TinyImageNet includes 10 tasks, each encompassing 20 classes. In this paper, we focus on class incremental learning, where the model cannot get the task index during testing. Our evaluation includes metrics following previous works~\cite{wu2024meta}. We choose the final average accuracy (Acc) across all tasks after sequential training on each task as the main metric for comparing approaches. Moreover, we use the averaged anytime accuracy (AAA) to evaluate the model throughout the stream of tasks. Let $AA_j$ denote the test average accuracy after training on $T_j$, then the evaluation metrics of AAA and Acc are:
\begin{equation}
AAA = \frac{1}{N}\sum_{j=1}^{N}AA_j, \quad Acc = AA_N.
\end{equation}

\begin{figure*}[t]
    \centering
    \includegraphics[width=1\textwidth]{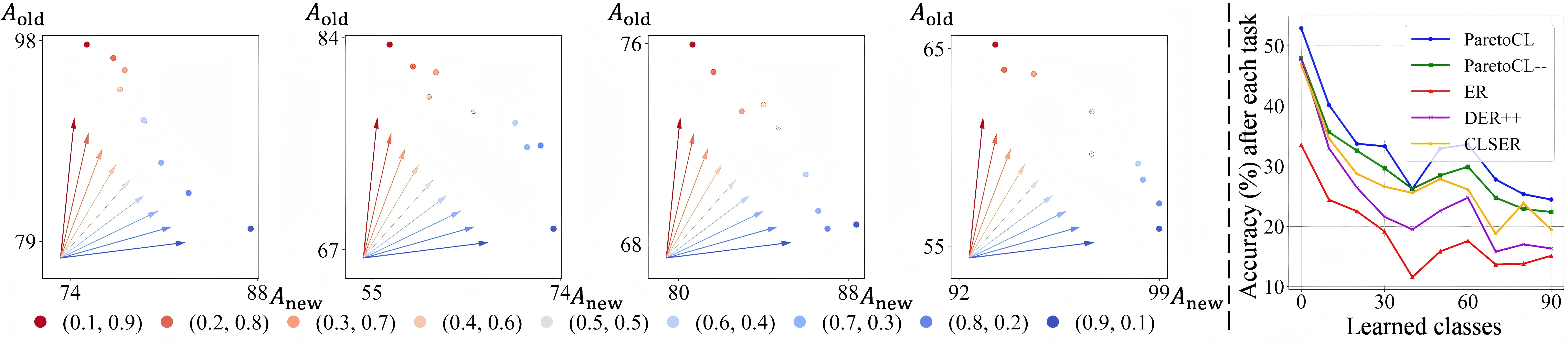}
    \caption{\textbf{Effectiveness of Preference-Conditioned Learning and Dynamic Preference Adaptation}. (left) shows the Pareto front approximated by ParetoCL in each stage on Seq-CIFAR10 in the offline setting. The results demonstrate that ParetoCL can effectively explore various stability-plasticity trade-offs; (right) compares the average incremental accuracy of different methods on Seq-CIFAR100 in the online setting. Dynamic Preference Adaptation can effectively improve performance compared to ParetoCL{-}{-} and other baselines.}
\label{fig:pareto_front}
\end{figure*}
\begin{table*}[t]
\centering
\resizebox{0.95\textwidth}{!}{%
\begin{tabular}{@{}l|cc|cc|cc@{}}
\toprule
\multirow{2}{*}{\textbf{Method}} & \multicolumn{2}{c|}{$\boldsymbol{M=600}$} & \multicolumn{2}{c|}{$\boldsymbol{M=1000}$} & \multicolumn{2}{c}{$\boldsymbol{M=1400}$} \\ \cline{2-7} 
                                 & AAA             & Acc            & AAA             & Acc            & AAA             & Acc             \\ \midrule
SGD                              & $11.63$\tiny{$\pm0.31$}  & $5.69$\tiny{$\pm0.19$}  & $11.63$\tiny{$\pm0.31$}  & $5.69$\tiny{$\pm0.19$}  & $11.63$\tiny{$\pm0.31$}  & $5.69$\tiny{$\pm0.19$}  \\ \midrule
OnEWC~\cite{Huszr2017NoteOT}     & $12.14$\tiny{$\pm0.43$}  & $6.23$\tiny{$\pm0.27$}  & $12.14$\tiny{$\pm0.43$}  & $6.23$\tiny{$\pm0.27$}  & $12.14$\tiny{$\pm0.43$}  & $6.23$\tiny{$\pm0.27$}  \\
LwF~\cite{Li2016LearningWF}      & $12.09$\tiny{$\pm0.66$}  & $6.32$\tiny{$\pm0.38$}  & $12.16$\tiny{$\pm0.49$}  & $5.99$\tiny{$\pm0.36$}  & $12.89$\tiny{$\pm0.21$}  & $6.35$\tiny{$\pm0.18$}  \\
IS~\cite{Zenke2017ContinualLT}   & $13.21$\tiny{$\pm0.59$}  & $7.33$\tiny{$\pm0.66$}  & $13.21$\tiny{$\pm0.59$}  & $7.33$\tiny{$\pm0.66$}  & $13.21$\tiny{$\pm0.59$}  & $7.33$\tiny{$\pm0.66$}  \\ \midrule
La-MAML~\cite{Gupta2020LaMAMLLM} & $10.42$\tiny{$\pm0.97$}  & $9.32$\tiny{$\pm0.48$}  & $13.45$\tiny{$\pm0.29$}  & $12.21$\tiny{$\pm0.47$} & $16.32$\tiny{$\pm0.41$}  & $15.52$\tiny{$\pm0.32$} \\
VR-MCL~\cite{wu2024meta}         & $25.31$\tiny{$\pm0.52$}  & $17.32$\tiny{$\pm0.48$} & $30.46$\tiny{$\pm0.49$}  & $22.31$\tiny{$\pm0.45$} & $32.22$\tiny{$\pm0.33$}  & $25.67$\tiny{$\pm0.40$} \\ \midrule
ER~\cite{Rolnick2018ExperienceRF}& $19.76$\tiny{$\pm0.31$}  & $12.69$\tiny{$\pm0.44$} & $18.70$\tiny{$\pm0.37$}  & $15.12$\tiny{$\pm0.76$} & $21.58$\tiny{$\pm0.42$}  & $18.25$\tiny{$\pm0.62$} \\
A-GEM~\cite{Chaudhry2018EfficientLL} & $14.60$\tiny{$\pm0.24$} & $6.59$\tiny{$\pm0.17$}  & $14.66$\tiny{$\pm0.32$}  & $6.24$\tiny{$\pm0.23$}  & $14.40$\tiny{$\pm0.28$}  & $6.75$\tiny{$\pm0.27$}  \\
GEM~\cite{LopezPaz2017GradientEM}& $16.97$\tiny{$\pm0.28$}  & $9.91$\tiny{$\pm0.39$}  & $16.23$\tiny{$\pm0.58$}  & $9.91$\tiny{$\pm0.32$}  & $16.03$\tiny{$\pm0.46$}  & $9.72$\tiny{$\pm0.41$}  \\
DER~\cite{buzzega2020dark}       & $14.62$\tiny{$\pm0.34$}  & $6.75$\tiny{$\pm0.19$}  & $14.65$\tiny{$\pm0.47$}  & $6.27$\tiny{$\pm0.13$}  & $6.65$\tiny{$\pm0.32$}   & $14.27$\tiny{$\pm0.44$} \\
DER++~\cite{buzzega2020dark}     & $24.15$\tiny{$\pm0.47$}  & $12.78$\tiny{$\pm0.52$} & $24.47$\tiny{$\pm0.55$}  & $16.32$\tiny{$\pm0.39$} & $22.55$\tiny{$\pm0.55$}  & $18.52$\tiny{$\pm0.51$} \\
CLSER~\cite{arani2021learning}   & $27.75$\tiny{$\pm0.42$}  & $16.97$\tiny{$\pm0.31$} & $27.83$\tiny{$\pm0.71$}  & $19.45$\tiny{$\pm0.35$} & $26.92$\tiny{$\pm0.57$}  & $24.96$\tiny{$\pm0.35$} \\
ER-OBC~\cite{Chrysakis2023OnlineBC} & $24.02$\tiny{$\pm0.34$} & $15.92$\tiny{$\pm0.39$} & $26.11$\tiny{$\pm0.54$}  & $17.31$\tiny{$\pm0.88$} & $28.91$\tiny{$\pm0.31$}  & $19.74$\tiny{$\pm0.83$} \\
OCM~\cite{Guo2022OnlineCL}      & $18.96$\tiny{$\pm0.22$}  & $8.72$\tiny{$\pm0.17$}  & $26.03$\tiny{$\pm0.90$}  & $15.88$\tiny{$\pm0.95$} & $28.23$\tiny{$\pm0.94$}  & $16.35$\tiny{$\pm0.81$} \\ 
\rowcolor[HTML]{EFEFEF} 
\textbf{ParetoCL (Ours)} & $\boldsymbol{31.03}$\tiny{$\boldsymbol{\pm0.61}$} & $\boldsymbol{20.49}$\tiny{$\boldsymbol{\pm0.29}$} & $\boldsymbol{33.04}$\tiny{$\boldsymbol{\pm0.61}$} & $\boldsymbol{24.45}$\tiny{$\boldsymbol{\pm0.39}$} & $\boldsymbol{33.76}$\tiny{$\boldsymbol{\pm0.39}$} & $\boldsymbol{26.97}$\tiny{$\boldsymbol{\pm0.42}$} \\ \bottomrule
\end{tabular}%
}
\caption{Ablation study on buffer size $M$ for Seq-CIFAR100 under online setting. The backbone is ResNet-18.}\label{tab:buffer-size-ablation}
\end{table*}

\noindent\textbf{Baseline and Training Details}
To effectively validate the efficacy of ParetoCL, we select three regularization-based methods (OnEWC~\cite{Huszr2017NoteOT}, SI~\cite{Zenke2017ContinualLT}, LwF~\cite{Li2016LearningWF}), two meta continual learning methods (La-MAML~\cite{Gupta2020LaMAMLLM}, VR-MCL~\cite{wu2024meta}), and six rehearsal-based methods (ER~\cite{Rolnick2018ExperienceRF}, GEM~\cite{LopezPaz2017GradientEM}, A-GEM~\cite{Chaudhry2018EfficientLL}, DER~\cite{buzzega2020dark}, DER++~\cite{buzzega2020dark}, CLSER~\cite{arani2021learning}, ER-OBC~\cite{Chrysakis2023OnlineBC}, OCM~\cite{Guo2022OnlineCL}). For fair comparison among different continual learning methods, we follow the hyperparameter settings used in their respective original papers and code implementations. We further provide a lower bound using SGD without any countermeasure to forgetting. For fair comparison, we train all models using the Stochastic Gradient Descent (SGD) optimizer. Following DER~\cite{buzzega2020dark}, we utilize ResNet-18 as the backbone. The hypernetwork $\Psi$ is an MLP with two hidden layers. The distribution for preferences during both training and inference is Dirichlet Distribution. The preferences sample number $K$ is 5 for training and 20 for inference. The learning rate of ParetoCL is set to 0.05 and the size of the replay buffer is set to 32 for all experiments. Under the offline CL setting, we set the training epochs to 5 for all methods. Further details are provided in supplementary material.

\subsection{Improvements over Baselines}
We conduct extensive baseline comparison on three datasets: Seq-CIFAR10, Seq-CIFAR100, and Seq-TinyImageNet with both online and offline settings and the results are summarized in Table \ref{tab:baseline_comp}. It is evident that our proposed ParetoCL significantly outperforms other replay-based methods (shown in the bottom row of the table), including methods with sophisticated replay strategies such as CLSER~\cite{arani2021learning}. These results align with our analysis in Figure \ref{fig:multistage} (right), demonstrating the effectiveness of learning a set of Pareto optimal models and dynamic adaption for different stability-plasticity trade-offs. Moreover, ParetoCL surpasses other representative state-of-the-art continual learning methods by a considerable margin. For instance, compared to VR-MCL~\cite{wu2024meta}, ParetoCL achieves an improvement of approximately \(6.6\%\), \(24.2\%\), and \(12.3\%\) in terms of final average accuracy on Seq-CIFAR10, Seq-CIFAR100, and Seq-TinyImageNet, respectively, under the online setting.

\subsection{Ablation Study}
\noindent\textbf{Effectiveness of preference-conditioned learning}
The primary motivation behind introducing Preference-Conditioned Learning is to efficiently learn a set of Pareto optimal models that embody different stability-plasticity trade-offs. To investigate whether ParetoCL successfully learns the mapping between preference vectors and corresponding models with different trade-offs, we conduct experiments on Seq-CIFAR10 in the offline setting. For each stage of the model, we use ten uniformly distributed preference vectors as input and evaluate the model's performance on the test sets of previous tasks and the current task. Figure \ref{fig:pareto_front} (left) illustrates the Pareto front approximated by ParetoCL. The well-distributed points along the Pareto front demonstrate that ParetoCL can effectively explore various stability-plasticity trade-offs. Interestingly, we observe that most solutions are clustered around the corner region of the Pareto front, suggesting that ParetoCL tends to favor balanced trade-offs that achieve good performance on both stability and plasticity objectives. This is a desirable property for CL systems, as overly prioritizing stability or plasticity can hinder the model's generalization ability.

\noindent\textbf{Effectiveness of dynamic preference adaptation}
The purpose of Dynamic Preference Adaptation is to adapt the stability-plasticity trade-off on the fly during inference. To verify its effectiveness, we design a variant termed ParetoCL{-}{-}, which employs Preference-Conditioned Learning during training but uses a static preference vector set at (0.5, 0.5) during inference. We conduct experiments on Seq-CIFAR100 to compare ParetoCL{-}{-} and ParetoCL, along with ER, CLSER, and DER++ under online setting. Figure \ref{fig:pareto_front}(right) visualizes the average incremental accuracy of different methods. We observe that Dynamic Preference Adaptation can effectively improve performance. It is worth noting that ParetoCL{-}{-} demonstrates competitive performance compared to other baselines, indicating that learning a set of models with different preferences during training allows the feature extractor to capture data distributions corresponding to various stability-plasticity trade-offs. The results also supports our view of generalization enhancement brought by objectives augmentation. On the other hand, ParetoCL{-}{-} can also serve as a simplified version of ParetoCL, reducing the computational cost of adapting to each sample.

\noindent\textbf{Impact of different memory buffer size}
In real-world applications, the storage cost of the memory buffer can vary significantly. For instance, storing samples on cloud servers is relatively easy, while it can be expensive on edge computing devices. Therefore, an ideal CL method should perform well under various buffer size settings. To investigate if ParetoCL maintains its performance across different memory buffer sizes, we conduct experiments on Seq-CIFAR100 in an online setting with varying $M$ as shown in Table~\ref{tab:buffer-size-ablation}. The results demonstrate that the performance of most methods improves as the memory buffer size $M$ increases. Notably, our ParetoCL consistently outperforms other baselines, suggesting that ParetoCL is effective and robust to different memory constraints. And we observe that GEM~\cite{LopezPaz2017GradientEM} and A-GEM~\cite{Chaudhry2018EfficientLL}, two gradient alignment replay methods, are insensitive to the buffer size, which is consistent with the result of ForkMerge~\cite{jiang2024forkmerge}.

\section{Discussion}
\label{sec:discussion}
In this section, we delve deeper into the characteristics and performance of ParetoCL by addressing two key questions. Further analysis can be found in supplementary material.

\noindent\textbf{How does ParetoCL's algorithm efficiency compare to other algorithms?}
Balancing computational cost and performance is a practical challenge in continual learning, particularly in online settings or resource-constrained environments~\cite{verwimp2023continual}. We analyze the time complexity of ParetoCL by comparing its training and testing time with other representative CL methods on the Seq-CIFAR10 dataset, as shown in Table~\ref{tab:training_time_comparison}. Despite achieving superior performance, ParetoCL requires less than one-fourth of the training time compared to the state-of-the-art method VR-MCL. As ParetoCL is an extension of the basic ER method, we also increased the training epochs of ER from 1 to 3 for a fair comparison. With nearly the same training time, ER achieves an accuracy of \(46.30\%\), while ParetoCL reaches \(59.95\%\), demonstrating that the performance gain is not solely due to multiple forward passes.

\noindent\textbf{Does learning a set of pareto optimal solutions lead to better generalization?}
As mentioned in the Introduction, our method can be viewed as a form of objective augmentation that enhances generalization. This hypothesis is preliminarily validated in Figure~\ref{fig:pareto_front}(right). To further verify this conjecture and exclude the influence of optimization, we directly applied multi-objective optimization methods MGDA~\cite{sener2018multi} and Tchebycheff Scalarization~\cite{choo1983proper}  to the plasticity and stability objectives ($\mathcal{L}_{replay}$ and $\mathcal{L}_{new}$). While these methods guarantee Pareto optimality, they do not simultaneously learn multiple objectives. Experiments on Seq-CIFAR10 and Seq-CIFAR100 demonstrate that ParetoCL outperforms these MOO variants, emphasizing the necessity of learning a set of models with different preferences.

\begin{table}[]
\centering

\resizebox{1\linewidth}{!}{
\begin{tabular}{l|ccc}
\toprule
\textbf{Method} & Time (s) & AAA & Acc \\
\midrule
GEM~\cite{LopezPaz2017GradientEM}   & 436.14 & 36.83 & 19.75 \\
La-MAML~\cite{Gupta2020LaMAMLLM}        & 693.42 & 36.76 & 35.44 \\
VR-MCL~\cite{wu2024meta}       & 1073.61 & 69.57 & 58.43 \\
DER++~\cite{buzzega2020dark}         & 108.81 & 60.63 & 50.33 \\
ER~\cite{Rolnick2018ExperienceRF}    & 79.21 & 52.74 & 33.14 \\
ER (3 epochs) & 288.71 & 59.21 & 46.30 \\
\rowcolor[HTML]{EFEFEF} 
\textbf{ParetoCL (Ours)} & 224.62 & 70.89 & 59.95 \\
\bottomrule
\end{tabular}
}
\caption{Training Time Comparison of Different Continual Learning Methods on Seq-CIFAR10.}\label{tab:training_time_comparison}
\setlength{\abovecaptionskip}{10pt} 
\end{table}

\begin{table}[]
\centering

\resizebox{0.99\linewidth}{!}{%
\begin{tabular}{l|ccc|ccc}
\toprule
\multirow{2}{*}{\textbf{Method}} & \multicolumn{3}{c|}{Seq-CIFAR10} & \multicolumn{3}{c}{Seq-CIFAR100} \\ \cline{2-7}
& AAA & Acc & \multicolumn{1}{c|}{\multirow{2}{*}{}} & AAA & Acc & \multicolumn{1}{c}{\multirow{2}{*}{}} \\
\midrule
MGDA~\cite{sener2018multi}                & 62.61 & 52.27 & & 24.23 & 16.54 & \\
Tchebycheff~\cite{choo1983proper}         & 52.03 & 46.71 & & 23.15 & 15.58 & \\
\rowcolor[HTML]{EFEFEF} 
\textbf{ParetoCL (Ours)}                                 & \textbf{70.89} & \textbf{59.95} & & \textbf{33.04} & \textbf{24.45} & \\
\bottomrule
\end{tabular}%
}
\caption{Comparison of methods on Seq-CIFAR10 and Seq-CIFAR100 datasets.}\label{tab:method-comparison}
\end{table}

\section{Conclusion}
In this work, we investigate the dynamic stability-plasticity trade-off in continual learning and reformulate it as a multi-objective optimization problem. We propose ParetoCL, a novel framework that efficiently learns a set of Pareto optimal solutions representing different trade-offs through preference-conditioned learning and enables dynamic adaptation during inference. Extensive experiments on multiple benchmarks demonstrate the superiority of ParetoCL over state-of-the-art continual learning methods.

\noindent\textbf{Limitations and Future Work.}
ParetoCL relies on a hypernetwork to generate the parameters of the preference-conditioned model. Exploring more sophisticated conditioning mechanisms may improve the performance. Additionally, applying the ParetoCL framework to other learning paradigms, such as continual reinforcement learning, is a promising direction for future research.

\section{Acknowledgments}
The work was supported by the Research Grants Council of the Hong Kong Special Administrative Region, China [GRF Project No. CityU 11215622], the National Natural Science Foundations of China (Grants No.62376267) and the innoHK project.

\bibliography{aaai25}

\end{document}